\documentclass[conference]{IEEEtran}
\IEEEoverridecommandlockouts
\usepackage{cite}
\usepackage{amsmath,amssymb,amsfonts}
\usepackage{algorithmic}
\usepackage{graphicx}
\usepackage{textcomp}
\usepackage{tabularx}
\usepackage{multicol}
\usepackage{multirow}
\usepackage{xcolor}
\def\BibTeX{{\rm B\kern-.05em{\sc i\kern-.025em b}\kern-.08em
    T\kern-.1667em\lower.7ex\hbox{E}\kern-.125emX}}
\begin{document}

\title{TC-GAT: Graph Attention Network for Temporal Causality Discovery}


\author{
    \IEEEauthorblockN{Xiaosong Yuan$^{a,b}$, Ke Chen$^{c}$, Wanli Zuo$^{a,b}$, Yijia Zhang$^{d}$\dag \thanks{\dag  Corresponding author: Yijia Zhang}}
    \IEEEauthorblockA{$^a$ College of Computer Science and Technology, Jilin University, Changchun, China}
    \IEEEauthorblockA{$^b$ Key Laboratory of Symbolic Computation and Knowledge Engineering, MOE, Changchun, China}
    \IEEEauthorblockA{$^c$ JD.com, Inc, Beijing, China}
    \IEEEauthorblockA{$^d$ College of Electronic Countermeasures, National University of Defense Technology, Hefei, China}
    \IEEEauthorblockA{\{yuanxs19,chenke19\}@mails.jlu.edu.cn, zuowl\_jlu@outlook.com, zhangyj\_gfkj@163.com}
}

\maketitle

\begin{abstract}
 The present study explores the intricacies of causal relationship extraction, a vital component in the pursuit of causality knowledge. Causality is frequently intertwined with temporal elements, as the progression from cause to effect is not instantaneous but rather ensconced in a temporal dimension. Thus, the extraction of temporal causality holds paramount significance in the field. In light of this, we propose a method for extracting causality from the text that integrates both temporal and causal relations, with a particular focus on the time aspect. To this end, we first compile a dataset that encompasses temporal relationships. Subsequently, we present a novel model, TC-GAT, which employs a graph attention mechanism to assign weights to the temporal relationships and leverages a causal knowledge graph to determine the adjacency matrix. Additionally, we implement an equilibrium mechanism to regulate the interplay between temporal and causal relations. Our experiments demonstrate that our proposed method significantly surpasses baseline models in the task of causality extraction.
\end{abstract}


\section{Introduction}
 The discovery of causality in natural language texts is a fundamental objective in computational linguistics, as a majority of causal relationships are expressed through natural language. The temporal aspects of causality are widely recognized to play a crucial role in accurately identifying cause-and-effect relationships between entities and events. It has been established that cause-and-effect pairs are frequently accompanied by time clues, where the cause precedes the effect and the co-occurrence of two events is a strong indicator of one being the cause of the other~\cite{zhang2021extracting, mirza2014analysis}. The extraction of temporal causality holds significant implications for various downstream tasks, such as dialogue creation and understanding, reading comprehension, and information extraction~\cite{ritter2010unsupervised,neji2016question,sun2018reading,huang2020semi}.

 In order to investigate the interplay between cause-effect relationships and time, we conduct an investigation into the extraction of causality from texts based on the temporal and causal characteristics of these relationships. Our approach involves the construction of a causal dataset that contains temporal relationships, which is then utilized by our novel model, TC-GAT. The TC-GAT model leverages graph attention to derive the weights of temporal relationships and utilizes the causal knowledge graph to obtain the adjacency matrix. Furthermore, we have developed an equilibrium mechanism to regulate the impact of temporal relationships on the extraction of causality. The results of our experiments demonstrate that our method yields substantial improvements in performance compared to baseline models on the causality extraction task.
 
 Table~\ref{tab1} provides two exemplars of temporal causal relationships, with the first example showcasing a simple time relationship between the cause (``\textbf{\emph{rain}}") and effect (``\textbf{\emph{floods}}"). On the other hand, the second example exhibits a more intricate set of temporal relationships. The cause (``\textbf{\emph{rain}}") is implicated in the occurrence of the effect (``\textbf{\emph{floods}}"). Due to the symmetry of time, the effect (``\textbf{\emph{floods}}") is reciprocally implicated in the cause (``\textbf{\emph{damaged}}" and ``\textbf{\emph{died}}"). Both of these effects are concurrent with the cause (``\textbf{\emph{rain}}") and effect (``\textbf{\emph{floods}}"), and are causally derived from the cause (``\textbf{\emph{rain}}").

 \begin{table}
 \caption{Simple and Complex Temporal relation in Causal relation.}\label{tab1}
 \centering
 \scalebox{1.0}{
 \begin{tabular}{|l|}
 \hline
 \textbf{Ex 1}: Simple Temporal relation in Causal relation\\
 The (e1:rain) caused the (e2:floods)\\
 \hline
 \textbf{Ex 2}: Complex Temporal relation in Causal relation\\
 Heavy (e1:rain) has been falling, causing (e2:floods), people's houses\\
 were (e3:damaged), and some people (e4:died) in the flood.\\
 \hline
 \end{tabular}}
 \end{table}

 To date, several models are proposed for causality extraction~\cite{hashimoto2014toward,riaz2014depth,luo2016commonsense}. However, the literature is characterized by a scarcity of models that are equipped to reason through the temporal relationships of events~\cite{mostafazadeh2016caters}. The latter study proposes a Comment Structure (CaTeRS) framework to facilitate judgments of the temporal relationships that are associated with causal relationships through the analysis of a substantial number of story-type events. A subsequent study, namely~\cite{ning2019joint}, introduced the Connection Structure TCR to infer the causality of events according to specified rules. However, these studies have yet to incorporate the use of deep learning models and rely heavily on the requirement for data annotation.

 Given this limitation, we present the TC-GAT model, which harnesses the power of deep neural networks to extract causal relationships through the examination of temporal relationships. The model integrates the Graph Attention GAT~\cite{velivckovic2017graph} to obtain attention weights of temporal relationships and concurrently constructs a knowledge graph acquisition adjacency matrix to predict causal events with greater accuracy.

 \begin{figure*}
   \centering
   \includegraphics[scale=0.5]{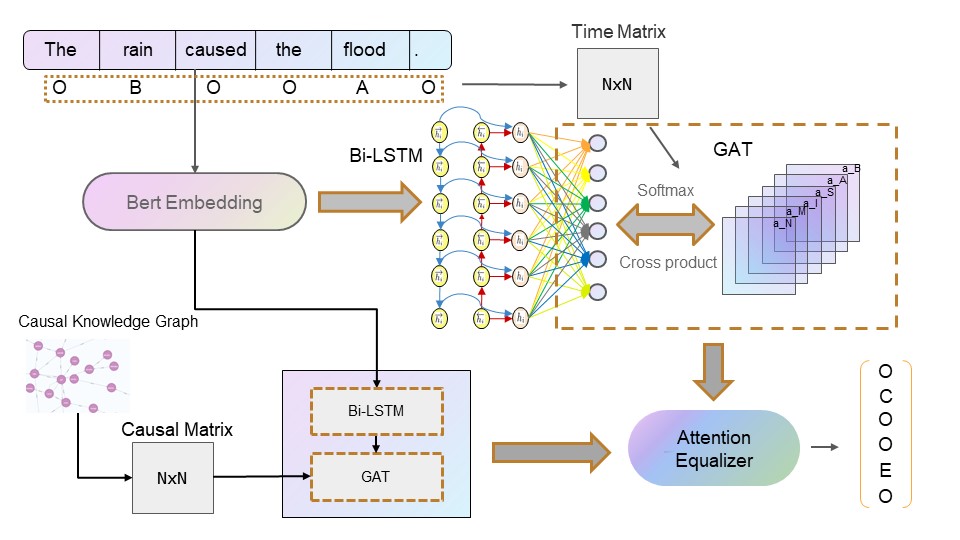}
   \caption{The overview of our approach, which consists of three major components:1) Based on temporal causal reasoning; 2) Based on the causal knowledge graph. 3) Based on the equilibrium mechanism.} \label{fig1}
 \end{figure*}

 The use of temporal information has the potential to intuitively enhance the extraction of causality. However, the conundrum lies in the fact that while causal entities must have a temporal relationship, they may not necessarily have a causal relationship. In light of this challenge, we propose an equilibrium mechanism that draws inspiration from~\cite{liuknowledge} to balance the impact of time information on the extraction of causality. Our experiments demonstrate a substantial improvement in the accuracy of predicted results.
 
 One of the central obstacles to discovering causality through temporal information is the need for joint labeled datasets. In this paper, we make the second contribution of annotating causal relation extraction datasets with temporal labels. We achieve this through two distinct methods. First, we annotate the causal data in the SemEval-2010 Task 8 dataset~\cite{hendrickx2019semeval} with temporal labels and use other relationships as negative samples to also annotate them with temporal labels. Second, we adopt a manual labeling approach to simultaneously annotate sentences with both temporal labels and causal labels in the Altlex dataset. In conclusion, this paper advances the state-of-the-art by presenting the following technical contributions: 
 \begin{itemize} 
  \item A thorough analysis and summary of the existing causality datasets, ultimately leading to the selection and revision of the SemEval2010-Task8 and Altlex datasets, which were then marked with causality and temporal relationship labels to form the TC-SemEval2010-Task8 and TC-Altlex temporal causal datasets.
  \item An analysis of temporal relationships, the construction of a causal knowledge graph, the integration of GAT to predict causal relations through temporal clues, and the proposal of a novel equilibrium mechanism to optimize the model, resulting in state-of-the-art performance.
  \item Extensive experiments to demonstrate the superiority of the proposed TC-GAT model.
 \end{itemize}

\section{Related Work}
 The goal of causal relation extraction is to uncover causality in texts automatically. Temporal causal relation extraction aims to identify causal relationships between events/entities that occur at different points in time, where the temporal order of the events is crucial to understanding the causal relationship.
 
 \subsection{Causality extraction}
 The task of causality extraction, which endeavors to discern the causal relationship between entities or events within texts, has elicited significant attention among scholars. Initial explorations of causality extraction focused on lexical and syntactic features, causal cues such as ``because" and ``result in", statistical features of event occurrence, and the construction of a term causality network from web-based text collections~\cite{hashimoto2014toward,gao2019modeling,riaz2014depth,luo2016commonsense,sil2010extracting,zhao2016event}. These methods primarily emphasized the computation of similarities in sentence syntax-dependent structure and the introduction of novel causal feature words. However, more recent research~\cite{kadowaki2019event} has leveraged the contextual relevance-enhancing capabilities of the BERT~\cite{devlin2018bert} to achieve remarkable results, despite neglecting the influence of temporal factors on entities or events.

 \subsection{Causality and Temporal joint extraction}
 Exploration for causality extraction through the lens of time relations garners considerable attention. Pioneering works such as Bethard~\cite{bethard2008learning} and Rink~\cite{rink2010learning} establish a close correlation between time relations and causality, with the subsequent proposal of the utilization of TimeML labeling~\cite{pustejovsky2003timeml} to annotate causal events. Unfortunately, they are limited by the progress of machine learning then. CATENA~\cite{mirza2016catena} adopts filtering rules grounded in time relations, while CAEVO~\cite{chambers2014dense} grows a temporal graph in a multi-sieve manner, where predictions are incrementally added through successive sieves, either needs to maintain a rules bank or updates a temporal graph constantly. CaTeRS~\cite{mostafazadeh2016caters} leverages a large corpus of story-type events to determine the time-related causality judgments, and TCR~\cite{ning2019joint} infers the causality of events based on specified rules.
 
 While these methods represent a range of approaches to causality extraction, many fall short in terms of their ability to harness the power of temporal relations for prediction or the capabilities of deep learning. This paper seeks to address this gap by presenting a novel approach based on joint temporal relationship reasoning, with a focus on utilizing existing classifications of time.

\section{Model}
 The overall architecture of our proposed model TC-GAT is depicted in Fig.~\ref{fig1}. Our method consists of 3 distinct yet interrelated major components. The first component entails the integration of temporal relations into the process of causality extraction. This is achieved through the deployment of graph attention, which serves to compute the joint weights of both temporal relations and textual features to enable the prediction of causal entities. The second component encompasses the utilization of a causal knowledge graph to obtain the adjacency matrix that is crucial for the graph attention network to effectively predict causal entities. Finally, the third component refers to the employment of an attention mechanism, which is based on the former equilibrium mechanism, to regulate the impact of temporal relations on causal relations. This mechanism plays a critical role in maintaining a dynamic balance between the two relations, ensuring the accuracy and robustness of the predictions.

 \subsection{BERT}
 The pre-trained language model BERT~\cite{devlin2018bert} has gained significant prominence in recent times for its demonstration of efficacy in comparison to traditional one-way language models and shallow splicing methods. The masked language model (MLM) approach utilized by BERT has been shown to provide more semantic information within generated language representations.

 Our work harnesses the capabilities of BERT by incorporating sentence vectors as inputs to the model, yielding contextual features that serve as word embeddings. The representation of each token in sentence S is obtained through the amalgamation of its respective token embedding, segment embedding, and position embedding. This is achieved by utilizing the sentence encoding of the form $[CLS, E_S, SEP]$, where CLS signifies the initiation of the sentence and SEP serves as an indicator of the sentence's conclusion.

 \begin{figure}[ht]
 \centering
 \includegraphics[scale=0.3]{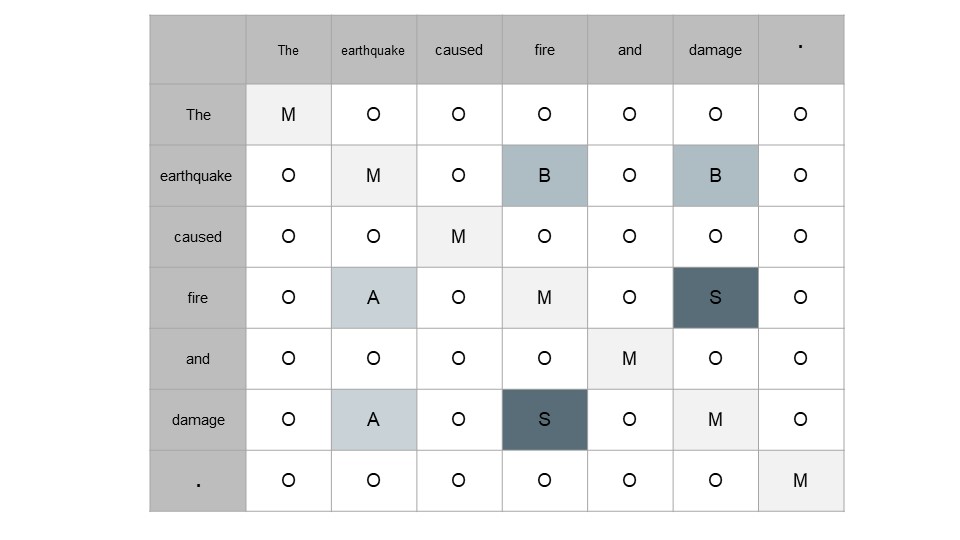}
 \caption{Time Matrix: is generated for the time tags in the sentence.} \label{fig2}
 \end{figure}

 \begin{figure}[ht]
 \centering
 \includegraphics[scale=0.4]{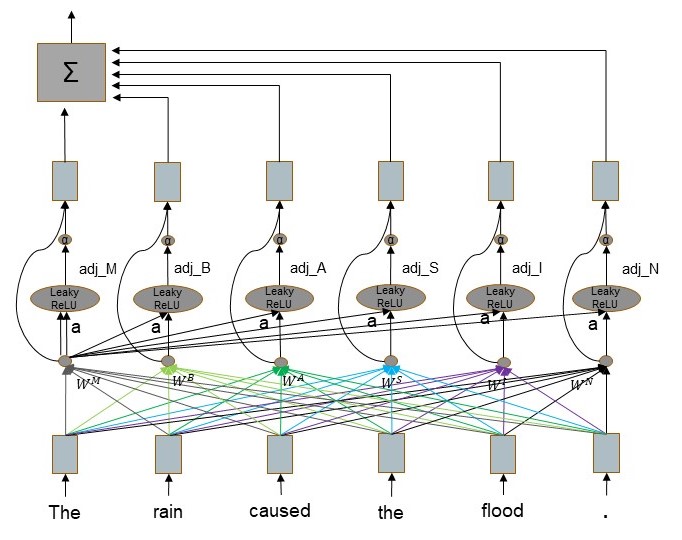}
 \caption{The T-GAT structure, omitting Bi-LSTM structure.} \label{fig3}
 \end{figure}
 
 \subsection{T-GAT}
 The BiLSTM model serves as the fundamental structure in our approach to extracting contextual information. By combining the outputs of this model with the Graph Attention Network (GAT) and a temporal relationship matrix, we are able to obtain feature vectors that encapsulate both contextual and temporal relationships. This synergistic utilization of the BiLSTM model and additional components enhances our ability to capture the intricacies of the data at hand, enabling the derivation of more nuanced insights.

 \subsubsection{BiLSTM}
 The Bi-directional Long Short-Term Memory network (BiLSTM) is a recurrent neural network architecture designed to capture contextual information from sequential input data by incorporating both past and future information. This is achieved by the integration of three gate structures, namely the input gate, the forget gate, and the output gate, which allow for selective forgetting of historical information, the addition of new information, and the integration of forward and backward information into the network's current state, respectively. The resulting contextual features, $h_{BiLSTM} = [\overrightarrow{h_t}, \overleftarrow{h_t}]$, are then employed in a multitude of natural language processing tasks, making the BiLSTM a ubiquitous presence in the field. In this work, we leverage the powerful text encoding capabilities of the BiLSTM by feeding word embeddings into the network, thereby obtaining rich and informative context features that are further utilized in our model.

 \subsubsection{GAT}
 We delve into the intricacies of employing a Graph Attention Network (GAT) to infer attention weights of temporal relationships. GAT, in essence, proposes a weighted summation of features of neighboring nodes via the utilization of an attention mechanism, enabling the determination of the significance of each neighboring node feature without being impacted by the graph structure. Unlike the Graph Convolutional Network (GCN) that employs fixed standardized operations, GAT replaces them with a neighbor node feature aggregation function that is guided by attention weights.

 To facilitate the integration of diverse temporal relationships into our model, we leverage GAT to obtain attention weights of temporal relationships. The first step is generating a time matrix for the time labels, where each word is annotated as M, and words without time relationships are marked as O. The different types of time relationships, namely, A (after), B (before), S (simultaneous), N (be-include), and I (include), are extracted to produce the corresponding time matrices $adj_A, adj_B, adj_S, adj_N, adj_I, adj_M$ as illustrated in Fig.\ref{fig2}.

 The T-GAT model structure is shown in Fig.\ref{fig3}: take each time state as a node to calculate the similarity coefficient, and take the M state as the neighbor node of each time state:
 \begin{small}
 \begin{equation}
 e_I=a ([hW_M,hW_I ])  ,I \in N=(B,A,S,I,M,N)    
 \end{equation}
 \end{small}
 Where $W_i \in R^{n \times m},i \in {B,A,S,I,M,N}$ is the weight matrix of each time state, $h \in R^{l \times n}$ is the context feature, $a \in R^{2l \times m}$ is the coefficient weight matrix.
 \begin{equation}
 \alpha_I = \frac{\exp(LeakyReLU(e_I))}{\sum_{k \in N}\exp (LeakyReLU(e_k))} \cdot adj_I  
 \end{equation}
 Where $\alpha_I$ is the obtained attention value, and $adj_I$ is the corresponding time matrix. According to the calculated attention coefficient, the weighted sum of features is as follows:
 \begin{equation}
 h_i^{'} = \sigma(\sum_{I \in N} \alpha_I \cdot W_I \cdot h)    
 \end{equation}
 At last, utilize multi-head attention to get the final result:
 \begin{equation}
 h_i^{'}(K) = \prod_{k=1}^K \sigma (\sum_{I \in N} \alpha_I^k \cdot W_I^k \cdot h)   
 \end{equation}

 \subsection{C-GAT}
 We present a novel approach that leverages the power of Graph Attention Networks (GAT) and the causal relationship adjacency matrix to construct a causal relationship-based graph attention network. This network, named C-GAT (Causal Graph Attention Networks), is based on the causal knowledge graph and can effectively capture the causal relationships in a sentence. We first input the word embedding vectors into a Bi-LSTM network to obtain the feature vectors, which encapsulate the contextual information of the words. Then, we integrate the causal relationship graph into the network by obtaining the adjacency matrix that reflects the causal relationships between the words. Finally, the feature vectors and the adjacency matrix are input into the GAT to obtain the output, which accurately captures the causal relationships.

 This innovative approach to modeling causal relationships provides significant advantages over conventional methods and contributes to a deeper understanding of the causal relationships in natural language processing tasks.

 \subsection{Equilibrium Mechanism}
 In this paper, we delve into the interplay between time and causality, specifically examining the implications of temporal relationships on causality extraction. Despite the axiomatic truth that the occurrence of entities must be grounded in the temporality of events, the presence of such relationships does not necessarily imply a causal relationship between them. As such, the indiscriminate consideration of temporal information can pose a significant hindrance to accurate causality assessment. To address this challenge, we introduce a novel equilibrium mechanism that endeavors to mitigate the pernicious influence of time relationships by learning context-sensitive attention weights. These weights are based on a number of trade-off factors that we consider, which are carefully designed to strike a balance between the utilization of temporal information and the preservation of causality judgment accuracy. The trade-off factors are as follows:
 
 \begin{equation}
 g = \alpha (W(h_i^{'} + h_{BERT} )+b)
 \end{equation}
 The trade-off coefficient \emph{g} is used to obtain the result vector after the trade-off:

 \begin{equation}
 h = gh_i^{'} + (1-g) h_{BERT}   
 \end{equation}
 Where $h_I^{'}$ and $H_BERT$ are the eigenvalues obtained by time reasoning and BERT, W, and b are the model parameters.

 \subsection{Model training and prediction}
 During training, we take the cross-entropy loss function as our loss function,
 \begin{equation}
 E = - \sum_{i=0}^n x_i \log P(x_i)
 \end{equation}
 Where $n = 2$, $x_i$ is the value of the \emph{i}-th dimension of the real label one hot form, $P(x_i)$ is the value of the \emph{i}-th dimension of the output prediction vector.

\section{Experiments}
 In this section, we present a comprehensive account of the time causal dataset by ourselves. Our contribution encompasses the nuanced classification of time, along with a meticulous description of the specific annotation techniques utilized. Moreover, we also undertake a comparative analysis of our proposed model and existing baseline methods, highlighting the unique characteristics and strengths of our approach. Furthermore, we provide a thorough examination of the results, offering insightful observations and revelations.

 \subsection{Experimental Setups}
 \subsubsection{Datasets and Evaluations}

 \begin{figure}[h]
 \centering
 \includegraphics[scale=0.45]{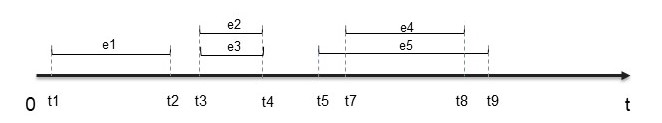}
 \caption{The sequence of events \textbf{\emph{e1, e2, e3, e4}} and \textbf{\emph{e5}} on the timeline.} \label{fig4}
 \end{figure}

 The temporal relationship constitutes an integral aspect of the entity's lifecycle and its influence on the latter can be clearly demarcated. In an effort to strike a balance between the number of categories and the performance of causality extraction, as well as to simplify the process of data labeling, we have classified the temporal relationship into five distinct categories: \textbf{Before(B)}, \textbf{After(A)}, \textbf{Include(I)}, \textbf{Be Include(N)}, and \textbf{Simultaneous(S)}.

 As demonstrated in Fig.\ref{fig4}, the temporal relationship between the five events \textbf{\emph{e1,e2,e3,e4}} and \textbf{\emph{e5}} can be illustrated in the following manner: event \textbf{\emph{e1}} occurs prior to events \textbf{\emph{e2,e3,e4}} and \textbf{\emph{e5}}, hence events \textbf{\emph{e2,e3,e4}} and \textbf{\emph{e5}} occur \textbf{After} event \textbf{\emph{e1}}. Furthermore, events \textbf{\emph{e2}} and \textbf{\emph{e3}} occur \textbf{Simultaneously} while events \textbf{\emph{e4}} and \textbf{\emph{e5}} fall under the \textbf{Include} and \textbf{Be-Included} categories, respectively. For the purpose of conciseness, any additional temporal relationships between the events are omitted.

 \begin{figure}
 \centering
 \includegraphics[trim={1.2cm 0.3cm 1.2cm 0.3cm},clip,scale=0.35]{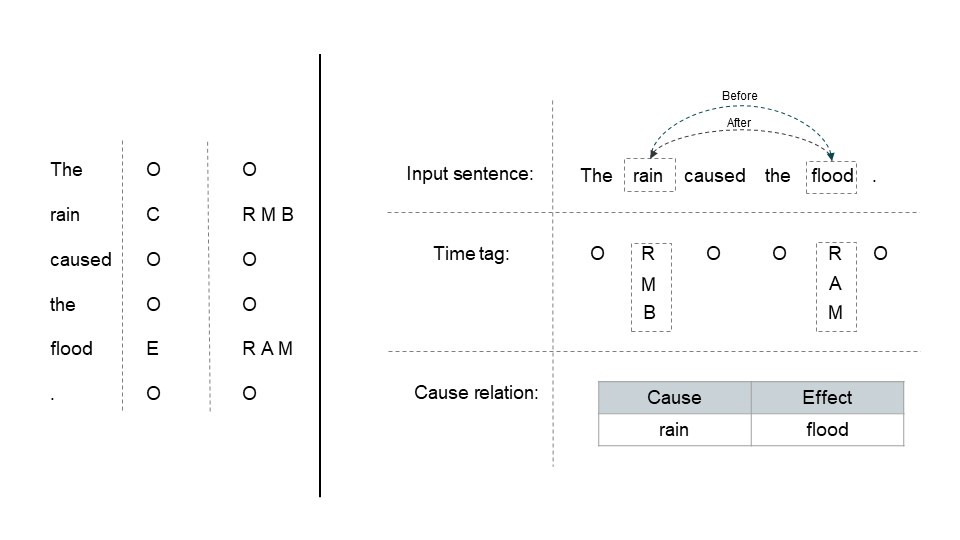}
 \caption{The data labeling method: \textbf{R} indicates that there is a time relationship, and \textbf{M} is used to represent itself, \textbf{B} and \textbf{A} respectively indicate the time relationship between event \textbf{\emph{rain}} and event \textbf{\emph{flood}}.} \label{fig5}
 \end{figure}

 In this study, the process of manually annotating a dataset that lacks temporal relationship labels in the field of causal relation extraction is described. Through a combination of a visual representation and a systematic approach to marking entities, the data labeling method was devised and implemented, as depicted in Fig.\ref{fig5}. The method involved marking each entity with a label ``R", followed by the relative temporal relationship with other entities and any additional relevant information. This manual annotation process is crucial to our research, as it allows for the introduction of temporal relationship labels into the field of causal relation extraction.

 \begin{table}[h]
	\caption{TC-SemEval2020-task8 and TC-Altlex datasets}  
	\label{tab:statistics}
	\centering
	\scalebox{1.1}{
		\begin{tabular}{ccc}
		  \hline
		  \bfseries Dataset &  \bfseries TC-SemEval2010-task8 &  \bfseries TC-AltLex \\
		  \hline
		   training & 2,094 & 867 \\
          test & 1,031 & 288 \\
          total & 3,125 & 1,155 \\
		  \hline
		\end{tabular}
	}
 \end{table}

 In our empirical investigation, we have subjected two datasets to our experimentation, both of which serve as a critical benchmark for assessing the performance of our proposed models. The first dataset, SemEval2010 task8~\cite{hendrickx2019semeval}, serves as a foundational corpus for our study, and its original composition includes 1,331 instances of causal relation data. To supplement and enhance the scope of our investigation, we have meticulously annotated an additional 1,794 instances of data with non-causal relations, endowing the SemEval2010 task8 corpus with a total of 3,125 annotated instances, with time labels included. The second dataset, TC-AltLex, represents a corpus of artificial annotations and has been derived from the 1,155 instances of data annotated with both causal and temporal relationships within the Altlex corpus~\cite{hidey2016identifying}. To evaluate the performance of our models, we have adopted the standard metrics of precision (P), recall (R), and F1-score (F1).

 \begin{table*}[h]
 \centering
 \small
 \caption{Experimental results under the coarse-grained dataset of TC-SemEval2010-Task8} 
 \label{tab:TC-SemEval2010-Task8}
 \resizebox{0.7\textwidth}{!}{
 \begin{tabular}{c|ccc|ccc|c}
 \hline
 \multirow{2}{*}{\textbf{Model}} & \multicolumn{3}{c|}{\textbf{C}} & \multicolumn{3}{c|}{\textbf{E}} & \multirow{2}{*}{\textbf{Macro-F1}} \\
 \cline{2-7}
 & P & R & F1 & P & R & F1 & \\
 \hline
 LSTM & 0.4888 & 0.3699 & 0.4212 & 0.5431 & 0.4385 & 0.4852 & 0.4532\\
 LSTM-CRF & 0.4824 & 0.4179 & 0.4479 & 0.6717 & 0.4806 & 0.5603 & 0.5041\\
 (char)LSTM-CRF & 0.4771 & 0.4399 & 0.4577 & 0.6300 & 0.5101 & 0.5637 & 0.5107\\
 BiLSTM & 0.6596 & 0.4223 & 0.5149 & 0.7133 & 0.5113 & 0.5957 & 0.5553\\
 BiLSTM-CRF & 0.5996 & 0.4504 & 0.5144 & 0.7478 & 0.5400 & 0.6272 & 0.5708\\
 BiLSTM-LAN & 0.7219 & 0.6633 & 0.6914 & 0.7604 & 0.6812 & 0.7186 & 0.7050\\
 Flair-BiLSTM-CRF & 0.7107 & 0.6193 & 0.6618 & 0.6570 & 0.6703 & 0.6636 & 0.6627\\
 BERT-softmax & 0.6745 & 0.4319 & 0.5266 & 0.7218 & 0.5416 & 0.6188 & 0.5727\\
 BERT-BiLSTM & 0.6937 & 0.5625 & 0.6213 & 0.7282 & 0.6893 & 0.7082 & 0.6648\\
 BERT-CRF & 0.7800 & 0.7259 & 0.7520 & 0.8024 & 0.7387 & 0.7692 & 0.7606\\
 C-GCN & 0.8619 & 0.7256 & 0.7879 & 0.7852 & 0.8398 & 0.8116 & 0.7998\\
 \hline
 TC-GAT & \textbf{0.9121} & \textbf{0.8966} & \textbf{0.9043} & \textbf{0.8831} & \textbf{0.9075} & \textbf{0.8951} & \textbf{0.8997}\\
 \hline
 \end{tabular}}
 \end{table*}
 
 \begin{table*}[h]
 \centering
 \small
 \caption{Experimental results under the coarse-grained dataset of TC-Altlex} 
 \label{tab:TC-Altlex}
 \resizebox{0.7\textwidth}{!}{
 \begin{tabular}{c|ccc|ccc|c}
 \hline
 \multirow{2}{*}{\textbf{Model}} & \multicolumn{3}{c|}{\textbf{C}} & \multicolumn{3}{c|}{\textbf{E}} & \multirow{2}{*}{\textbf{Macro-F1}} \\
 \cline{2-7}
 & P & R & F1 & P & R & F1 & \\
 \hline
 LSTM & 0.4167 & 0.3383 & 0.3734 & 0.4237 & 0.3534 & 0.3854 & 0.3794\\
 LSTM-CRF & 0.4766 & 0.3343 & 0.3930 & 0.4512 & 0.4000 & 0.4241 & 0.4086\\
 (char)LSTM-CRF & 0.4752 & 0.3433 & 0.3986 & 0.4688 & 0.4030 & 0.4334 & 0.4160\\
 BiLSTM &0.4978 & 0.3455 & 0.4079 & 0.5279 & 0.3494 & 0.4205 & 0.4142\\
 BiLSTM-CRF & 0.5614 & 0.3542 & 0.4344 & 0.5440 & 0.3646 & 0.4366 & 0.4355\\
 BiLSTM-LAN & 0.5000 & 0.4755 & 0.4874 & 0.5523 & 0.4567 & 0.5000 & 0.4937\\
 Flair-BiLSTM-CRF & 0.5182 & 0.3497 & 0.4176 & 0.5907 & 0.4955 & 0.5390 & 0.4783\\
 BERT-softmax & 0.5484 & 0.5113 & 0.5292 & 0.5246 & 0.6784 & 0.5917 & 0.5605\\
 BERT-BiLSTM & 0.5752 & 0.4435 & 0.5008 & 0.5928 & 0.5014 & 0.5433 & 0.5221\\
 BERT-CRF & 0.6261 & 0.6779 & 0.6510 & 0.6477 & 0.6806 & 0.6638 & 0.6574\\
 C-GCN & 0.7270 & 0.7406 & 0.7337 & 0.7321 & 0.8693 & 0.7948 & 0.7643\\
 \hline
 TC-GAT & \textbf{0.8707} & \textbf{0.9182} & \textbf{0.8938} & \textbf{0.9255} & \textbf{0.9255} & \textbf{0.9255} & \textbf{0.9097}\\
 \hline
 \end{tabular}}
 \end{table*}
 
 \subsubsection{Implementation Details}
 In our experimental setting, the text length is constrained to a maximum of 50 units, while the batch size is set to 24. Furthermore, the word embedding layer is endowed with a 300-dimensional(d) representation for each word in the corpus. The learning rate (lr) for our model is 1e-3, while the lr for the BERT module is 1e-5. The BiLSTM layer comprises a hidden layer with 150 units, and the T-GAT layer is 100-d, with a multi-head mechanism consisting of 3 heads and a dropout value of 0.15, and an alpha value of 0.008. In a similar vein, the C-GAT layer is specified to 100-d, a multi-head mechanism with 3 heads, and a dropout of 0.15, along with an alpha value of 0.008. The BERT layer is a 768-d representation with a dropout value of 0.1.

 \begin{table*}[ht]
 \caption{TC-GAT model ablation study} \label{tab:AblationStudy}
 \begin{center}
 \resizebox{0.7\textwidth}{!}{
 \begin{tabular}{lccccccc}
 \hline
 \multirow{2}{*}{\textbf{Methods}} & \multicolumn{7}{c}{\textbf{Metrics}} \\
 \cline{2-8}
 & C-P & C-R & C-F1 & E-P & E-R & E-F1 & Macro-F1 \\
 \hline
 \multicolumn{5}{l}{Train set: SemEval2010-task8}  \\
 \hline
 TC-GAT & 0.9121 & 0.8966 & 0.9043 & 0.8831 & \textbf{0.9075} & \textbf{0.8951} & \textbf{0.8997} \\
 NO-BERT & 0.8942 & 0.8682 & 0.8810 & 0.8667 & 0.8625 & 0.8646 & 0.8728 \\
 NO-Equilibrium & \textbf{0.9507} & 0.8921 & \textbf{0.9205} & 0.8687 & 0.8883 & 0.8784 & 0.8995\\
 T-GAT & 0.8752 & \textbf{0.9007} & 0.8878 & 0.8676 & 0.8803 & 0.8739 & 0.8809\\
 C-GAT & 0.8382 & 0.8784 & 0.8579 & 0.8569 & 0.8625 & 0.8597 & 0.8588\\
 BERT & 0.6745 & 0.4319 & 0.5266 & 0.7218 & 0.5416 & 0.6188 & 0.5727\\
 \hline
 \multicolumn{5}{l}{Train set: TC-AltLex} \\
 \hline
 TC-GAT & 0.8707 & \textbf{0.9182} & \textbf{0.8938} & \textbf{0.9255} & \textbf{0.9255} & \textbf{0.9255 } & \textbf{0.9097} \\
 NO-BERT & 0.8589 & 0.8303 & 0.8444 & 0.9072 & 0.8682 & 0.8873 & 0.8659 \\
 NO-Equilibrium & \textbf{0.9338} & 0.8545 & 0.8924 & 0.8817 & 0.8968 & 0.8892 & 0.8908\\
 T-GAT & 0.8459 & 0.8485 & 0.8472 & 0.9127 & 0.8682 & 0.8899 & 0.8686\\
 C-GAT & 0.8383 & 0.8485 & 0.8434 & 0.8300 & 0.8395 & 0.8348 & 0.8391\\
 BERT & 0.5484 & 0.5113 & 0.5292 & 0.5246 & 0.6784 & 0.5917 & 0.5605\\
 \hline
 \end{tabular}}
 \end{center}
 \end{table*}
 
 \subsubsection{Baselines}
 In this investigative endeavor, we engage in a thorough examination of a range of baseline models through a battery of experiments conducted on two datasets. The compared models encompass a spectrum of both conventional and cutting-edge neural architectures, including:
 \begin{itemize}
     \item \textbf{LSTM} The Long Short-Term Memory model~\cite{hochreiter1997long} leverages a gated structure to learn long-term dependencies.
     \item \textbf{LSTM-CRF} The LSTM-CRF model of~\cite{augustyniak2019aspect} extends the LSTM model by incorporating a CRF layer to take into account correlations between class labels.
     \item \textbf{(char)LSTM-CRF} The character-enhanced LSTM-CRF model of~\cite{augustyniak2019aspect} combines word embeddings with character-based representations, thereby providing greater expressive power for the neural architecture to realize more open and informative representations.
     \item \textbf{BiLSTM} The Bidirectional Long Short-Term Memory (BiLSTM) model~\cite{zhang2015bidirectional}, which is based on the LSTM model, but integrates both the forward and backward inputs of a sequence.
     \item \textbf{BiLSTM-CRF} This model~\cite{huang2015bidirectional} incorporates contextual information and adds a CRF layer to consider correlations between category labels.
     \item \textbf{BiLSTM-LAN} The Hierarchically Refined Label Attention Network (BiLSTM-LAN) model of~\cite{cui2019hierarchically} utilizes label embeddings to capture latent, long-term dependencies by incrementally refining label distributions through hierarchical attention.
     \item \textbf{Flair-BiLSTM-CRF} The Contextual String Embeddings~\cite{akbik2018contextual} employs the internal state of a pre-trained character language model to generate a novel type of word embedding, modeling words as sequences of characters, and providing diverse embeddings for the same word.
     \item \textbf{BERT} The Bidirectional Encoder Representations from Transformers (BERT) model of~\cite{devlin2018bert}, which harnesses the Transformer structure to yield highly effective features.
     \item \textbf{BERT-LSTM} The BERT-BiLSTM model employs a pre-trained BERT for embedding and BiLSTM to acquire contextual features.
     \item \textbf{BERT-CRF} The BERT-CRF model of~\cite{souza2019portuguese} blends the transfer capability of BERT with the structured prediction capabilities of CRF.
     \item \textbf{C-GCN} The Convolutional Graph Convolutional Network (C-GCN) model of~\cite{zhang2018graph} leverages a BiLSTM module as a context encoder and employs GCN to transform temporal relationships.
 \end{itemize}

 \renewcommand{\thetable}{7}
 \begin{table*}[h]
 \centering
 \caption{Experimental results under the SemEval2010-task8 dataset} 
 \label{tab:SemEvalallresult}
 \resizebox{0.7\textwidth}{!}{
 \begin{tabular}{c|ccc|ccc|c}
 \hline
 \multirow{2}{*}{\textbf{Model}} & \multicolumn{3}{c|}{\textbf{C}} & \multicolumn{3}{c|}{\textbf{E}} & \multirow{2}{*}{\textbf{Macro-F1}} \\
 \cline{2-7}
 & P & R & F1 & P & R & F1 & \\
 \hline
 (char)BiLSTM-CRF & 0.6858 & 0.6248 & 0.6538 & 0.7495 & 0.7160 & 0.7324 & 0.6931\\
 BiLSTM-LAN & 0.7252 & 0.6634 & 0.6929 & 0.7632 & 0.7323 & 0.7474 & 0.7202\\
 Flair-BiLSTM-CRF & 0.7460 & 0.5455 & 0.6302 & \textbf{0.8253} & 0.6227 & 0.7098 & 0.6700\\
 BERT-CRF & 0.6938 & \textbf{0.7408} & 0.7166 & 0.7457 & \textbf{0.7850} & \textbf{0.7648} & 0.7407\\
 C-GAT & \textbf{0.7839} & 0.7198 & \textbf{0.7505} & 0.7728 & 0.7413 & 0.7568 & \textbf{0.7537}\\
 \hline
 \end{tabular}}
 \end{table*}

 \renewcommand{\thetable}{6}
 \begin{table}[h]
	\caption{SemEval2020-task8 total statistics}  
	\label{tab:SemEvalall}
	\centering
	\scalebox{1.1}{
		\begin{tabular}{cccc}
		  \hline
		       Dataset & training & test & total \\
		  \hline
		       SemEval2010-task8 & 7,135 & 3,514 & 10,649 \\
		  \hline
		\end{tabular}
	}
\end{table}

 \subsection{Experimental Results}
 The coarse-grained experimental results of the TC-SemEval2010-Task8 dataset are shown in Table~\ref{tab:TC-SemEval2010-Task8}, and Table~\ref{tab:TC-Altlex} shows the results of the TC-Altlex dataset, the Macro-F1 values of the TC-GAT in the two datasets are 89.97\% and 90.97\%, respectively, and the accuracy is also the highest, which is better than others.

 The empirical results of this study demonstrate that the inclusion of temporal relationships within both the C-GCN model and the TC-GAT model proposed herein has led to a significant improvement in the extraction of causal relationships. This attests to the viability of this line of inquiry. On the one hand, the TC-GAT model leverages graph attention for temporal relationships to concentrate on words imbued with temporal significance, and on the other hand, it employs a causal relationship knowledge graph to focus on words with a potential causal connotation, further augmented by the integration of both BERT and BiLSTM to capture the contextual information of words. Such synergistic combination engenders a more robust extraction of causal relationships. Furthermore, the comparison between the C-GCN model and the TC-GAT model reveals that the implementation of an equalization mechanism effectively counteracts the adverse effects associated with the temporal relationships.

 \subsection{Ablation Study}
 To maintain a level of impartiality, all hyper-parameters are held constant while the innovative additions are integrated and retrained. The proposed model employs BERT as an embedding encoder of contextual features, with the incorporation of T-GAT to calculate the temporal attention weight, and C-GAT to compute the causal attention weight. The final addition of an equalization mechanism, referred to as Equilibrium, serves to balance the impact of the temporal relationship, culminating in the complete expression of the methodology.

 In the ablation experiment, Table ~\ref{tab:AblationStudy} demonstrates the results of evaluating each constituent module's efficacy, yielding positive outcomes on both datasets. This confirms the validity of the proposed method across both datasets. In the first experiment, the removal of BERT as the embedding encoding context feature was tested, resulting in a decrease of 2.69\% and 4.38\% in Macro-F1 value in the two datasets, respectively, thereby attesting to BERT's efficacy in capturing contextual information. Subsequently, the removal of the equalization mechanism based on the original model resulted in a relatively modest drop in performance by 0.02\% and 1.89\%, however, instability in prediction was observed when comparing the results of labels C and E, thus highlighting the importance of the balance mechanism in equilibrating the weight outcomes of T-GAT and C-GAT. The performance of T-GAT and C-GAT models was then evaluated in the two datasets, yielding improved performance, thereby affirming the time relationship's ability to enhance causal relationship extraction accuracy, as well as the impact of the causal knowledge graph. However, it is important to note that the performance of these modules is inferior to the overall model, and the disparity between the P-value and R-value in the T-GAT model highlights the deleterious effect of the temporal relationship.

 The generalization capabilities of the C-GAT module of the proposed model were evaluated through an exploratory investigation. To this end, the entire dataset of SemEval2010-Task8 was selected for comparison, taking into consideration the constraint imposed on the longest text of 50 words and the consequent removal of sentences exceeding this limit. The statistics of the retained sentences are presented in Table~\ref{tab:SemEvalall}.

 The comparative evaluation of the C-GAT model proposed in this study against Flair, BiLSTM-LAN, BERT-CRF, and (Char)BiLSTM-CRF, reveals the superiority of the former in terms of coarse granularity, as exhibited in Table~\ref{tab:SemEvalallresult}. The Macro-F1 value of the C-GAT model stands at 75.37\%, surpassing that of the BERT-CRF model, which indicates the efficacy of BERT in representing contextual and semantic features of text and the capacity of CRF in incorporating inter-label relationships into the model. The utilization of multiple embeddings in Flair contributes to the refinement of semantic features, while the hierarchical attention per word in the LAN module enables the capture of long-term label dependencies.

\section{Conclusion and Future Work}
 In conclusion, this paper presents a novel TC-GAT model that leverages the graph attention mechanism to effectively capture both temporal and causal relationships in the text. The addition of a balance mechanism serves to equilibrate the influence of temporal factors on causality, resulting in improved performance as demonstrated through comparison with established baselines. Future research directions include the integration of entity boundary prediction processing, as well as the exploration of complementary models from the state-of-the-art to further enhance performance in datasets lacking time labels.

\section*{Acknowledgment}
 This work is supported by the National Natural Science Foundation of China under Grant (61976103), and the general foundation of the National University of Defense Technology under Grant (ZK22-11).








\begin{thebibliography}{00}









\bibitem{mirza2014analysis}
P~Mirza and S~Tonelli, ``An analysis of causality between events and its
  relation to temporal information,'' in \emph{Proceedings of COLING 2014, the
  25th International Conference on Computational Linguistics: Technical
  Papers}, 2014, pp. 2097--2106.

\bibitem{velivckovic2017graph}
P~Veli{\v{c}}kovi{\'c}, G~Cucurull, A~Casanova, A~Romero, P~Lio, and
  Y~Bengio, ``Graph attention networks,'' \emph{arXiv preprint
  arXiv:1710.10903}, 2017.

\bibitem{hendrickx2019semeval}
I~Hendrickx, S~N. Kim, Z~Kozareva, P~Nakov, D~O. S{\'e}aghdha,
  S~Pad{\'o}, M~Pennacchiotti, L~Romano, and S~Szpakowicz, ``Semeval-2010
  task 8: Multi-way classification of semantic relations between pairs of
  nominals,'' \emph{arXiv preprint arXiv:1911.10422}, 2019.

\bibitem{liuknowledge}
J~Liu, Y~Chen, and J~Zhao, ``Knowledge enhanced event causality
  identification with mention masking generalizations.''

\bibitem{hashimoto2014toward}
C~Hashimoto, K~Torisawa, J~Kloetzer, M~Sano, I~Varga, J.-H. Oh, and
  Y~Kidawara, ``Toward future scenario generation: Extracting event causality
  exploiting semantic relation, context, and association features,'' in
  \emph{Proceedings of the 52nd Annual Meeting of the Association for
  Computational Linguistics (Volume 1: Long Papers)}, 2014, pp. 987--997.

\bibitem{gao2019modeling}
L~Gao, P~K. Choubey, and R~Huang, ``Modeling document-level causal
  structures for event causal relation identification,'' in \emph{Proceedings
  of the 2019 Conference of the North American Chapter of the Association for
  Computational Linguistics: Human Language Technologies, Volume 1 (Long and
  Short Papers)}, 2019, pp. 1808--1817.

\bibitem{riaz2014depth}
M~Riaz and R~Girju, ``In-depth exploitation of noun and verb semantics to
  identify causation in verb-noun pairs,'' in \emph{Proceedings of the 15th
  Annual Meeting of the Special Interest Group on Discourse and Dialogue
  (SIGDIAL)}, 2014, pp. 161--170.

\bibitem{hu2017inference}
Z~Hu, E~Rahimtoroghi, and M~A. Walker, ``Inference of fine-grained event
  causality from blogs and films,'' \emph{arXiv preprint arXiv:1708.09453},
  2017.

\bibitem{luo2016commonsense}
Z~Luo, Y~Sha, K~Q. Zhu, S.-w. Hwang, and Z~Wang, ``Commonsense causal
  reasoning between short texts.'' in \emph{KR}, 2016, pp. 421--431.

\bibitem{kadowaki2019event}
K~Kadowaki, R~Iida, K~Torisawa, J.-H. Oh, and J~Kloetzer, ``Event causality
  recognition exploiting multiple annotators’ judgments and background
  knowledge,'' in \emph{Proceedings of the 2019 Conference on Empirical Methods
  in Natural Language Processing and the 9th International Joint Conference on
  Natural Language Processing (EMNLP-IJCNLP)}, 2019, pp. 5820--5826.

\bibitem{devlin2018bert}
J~Devlin, M.-W. Chang, K~Lee, and K~Toutanova, ``Bert: Pre-training of deep
  bidirectional transformers for language understanding,'' \emph{arXiv preprint
  arXiv:1810.04805}, 2018.

\bibitem{bethard2008learning}
S~Bethard and J~H. Martin, ``Learning semantic links from a corpus of
  parallel temporal and causal relations,'' in \emph{Proceedings of ACL-08:
  HLT, Short Papers}, 2008, pp. 177--180.

\bibitem{rink2010learning}
B~Rink, C~A. Bejan, and S~Harabagiu, ``Learning textual graph patterns to
  detect causal event relations,'' in \emph{Twenty-Third International FLAIRS
  Conference}, 2010.

\bibitem{pustejovsky2003timeml}
J~Pustejovsky, J~M. Castano, R~Ingria, R~Sauri, R~J. Gaizauskas,
  A~Setzer, G~Katz, and D~R. Radev, ``Timeml: Robust specification of event
  and temporal expressions in text.'' \emph{New directions in question
  answering}, vol~3, pp. 28--34, 2003.

\bibitem{mirza2016catena}
P~Mirza and S~Tonelli, ``Catena: Causal and temporal relation extraction from
  natural language texts,'' in \emph{Proceedings of COLING 2016, the 26th
  International Conference on Computational Linguistics: Technical Papers},
  2016, pp. 64--75.

\bibitem{mostafazadeh2016caters}
N~Mostafazadeh, A~Grealish, N~Chambers, J~Allen, and L~Vanderwende,
  ``Caters: Causal and temporal relation scheme for semantic annotation of
  event structures,'' in \emph{Proceedings of the Fourth Workshop on Events},
  2016, pp. 51--61.

\bibitem{ning2019joint}
Q~Ning, Z~Feng, H~Wu, and D~Roth, ``Joint reasoning for temporal and causal
  relations,'' \emph{arXiv preprint arXiv:1906.04941}, 2019.

\bibitem{sil2010extracting}
A~Sil, F~Huang, and A~Yates, ``Extracting action and event semantics from
  web text.'' in \emph{AAAI Fall Symposium: Commonsense Knowledge}.\hskip 1em
  plus 0.5em minus 0.4em\relax Citeseer, 2010.

\bibitem{zhao2016event}
S~Zhao, T~Liu, S~Zhao, Y~Chen, and J.-Y. Nie, ``Event causality extraction
  based on connectives analysis,'' \emph{Neurocomputing}, vol. 173, pp.
  1943--1950, 2016.

\bibitem{kingma2015adam}
D~P. Kingma and J~Ba, ``Adam: A methodfor stochastic optimization,'' in
  \emph{International Conference onLearning Representations (ICLR)}, 2015.

\bibitem{hochreiter1997long}
S~Hochreiter and J~Schmidhuber, ``Long short-term memory,'' \emph{Neural
  computation}, vol~9, no~8, pp. 1735--1780, 1997.

\bibitem{huang2015bidirectional}
Z~Huang, W~Xu, and K~Yu, ``Bidirectional lstm-crf models for sequence
  tagging,'' \emph{arXiv preprint arXiv:1508.01991}, 2015.

\bibitem{zhang2015bidirectional}
S~Zhang, D~Zheng, X~Hu, and M~Yang, ``Bidirectional long short-term memory
  networks for relation classification,'' in \emph{Proceedings of the 29th
  Pacific Asia conference on language, information and computation}, 2015, pp.
  73--78.

\bibitem{zhang2021extracting}
S~Zhang, L~Huang, and Q~Ning, ``Extracting temporal event relation with
  syntactic-guided temporal graph transformer,'' \emph{arXiv preprint
  arXiv:2104.09570}, 2021.

\bibitem{ritter2010unsupervised}
A~Ritter, C~Cherry, and W~B. Dolan, ``Unsupervised modeling of twitter
  conversations,'' in \emph{Human Language Technologies: The 2010 Annual
  Conference of the North American Chapter of the Association for Computational
  Linguistics}, 2010, pp. 172--180.

\bibitem{neji2016question}
Z~Neji, M~Ellouze, and L~H. Belguith, ``Question answering based on temporal
  inference.'' \emph{Res. Comput. Sci.}, vol. 117, pp. 133--141, 2016.

\bibitem{sun2018reading}
Y~Sun, G~Cheng, and Y~Qu, ``Reading comprehension with graph-based
  temporal-casual reasoning,'' in \emph{Proceedings of the 27th International
  Conference on Computational Linguistics}, 2018, pp. 806--817.

\bibitem{huang2020semi}
L~Huang and H~Ji, ``Semi-supervised new event type induction and event
  detection,'' in \emph{Proceedings of the 2020 Conference on Empirical Methods
  in Natural Language Processing (EMNLP)}, 2020, pp. 718--724.

\bibitem{chambers2014dense}
N~Chambers, T~Cassidy, B~McDowell, and S~Bethard, ``Dense event ordering
  with a multi-pass architecture,'' \emph{Transactions of the Association for
  Computational Linguistics}, vol~2, pp. 273--284, 2014.

\bibitem{hidey2016identifying}
C~Hidey and K~McKeown, ``Identifying causal relations using parallel
  wikipedia articles,'' in \emph{Proceedings of the 54th Annual Meeting of the
  Association for Computational Linguistics (Volume 1: Long Papers)}, 2016, pp.
  1424--1433.

\bibitem{zhang2018graph}
Y~Zhang, P~Qi, and C~D. Manning, ``Graph convolution over pruned dependency
  trees improves relation extraction,'' \emph{arXiv preprint arXiv:1809.10185},
  2018.

\bibitem{cui2019hierarchically}
L~Cui and Y~Zhang, ``Hierarchically-refined label attention network for
  sequence labeling,'' \emph{arXiv preprint arXiv:1908.08676}, 2019.

\bibitem{akbik2018contextual}
A~Akbik, D~Blythe, and R~Vollgraf, ``Contextual string embeddings for
  sequence labeling,'' in \emph{Proceedings of the 27th international
  conference on computational linguistics}, 2018, pp. 1638--1649.

\bibitem{dunietz2017because}
J~Dunietz, L~Levin, and J~G. Carbonell, ``The because corpus 2.0: Annotating
  causality and overlapping relations,'' in \emph{Proceedings of the 11th
  Linguistic Annotation Workshop}, 2017, pp. 95--104.

\bibitem{augustyniak2019aspect}
{\L}~Augustyniak, T~Kajdanowicz, and P~Kazienko, ``Aspect detection using
  word and char embeddings with (bi) lstm and crf,'' in \emph{2019 IEEE second
  international conference on artificial intelligence and knowledge engineering
  (AIKE)}.\hskip 1em plus 0.5em minus 0.4em\relax IEEE, 2019, pp. 43--50.

\bibitem{souza2019portuguese}
F~Souza, R~Nogueira, and R~Lotufo, ``Portuguese named entity recognition
  using bert-crf,'' \emph{arXiv preprint arXiv:1909.10649}, 2019.
\end{thebibliography}
\end{document}